# A Data Mining Approach to Flight Arrival Delay Prediction for American Airlines

Navoneel Chakrabarty
*Computer Science and Engineering*
*Jalpaiguri Government Engineering College*
Jalpaiguri, West Bengal, India
nc2012@cse.jgec.ac.in

*Abstract*—In the present scenario of domestic flights in USA, there have been numerous instances of flight delays and cancellations. In the United States, the American Airlines, Inc. have been one of the most entrusted and the world's largest airline in terms of number of destinations served. But when it comes to domestic flights, AA has not lived up to the expectations in terms of punctuality or on-time performance. Flight Delays also result in airline companies operating commercial flights to incur huge losses. So, they are trying their best to prevent or avoid Flight Delays and Cancellations by taking certain measures. This study aims at analyzing flight information of US domestic flights operated by American Airlines, covering top 5 busiest airports of US and predicting possible arrival delay of the flight using Data Mining and Machine Learning Approaches. The Gradient Boosting Classifier Model is deployed by training and hyper-parameter tuning it, achieving a maximum accuracy of 85.73%. Such an Intelligent System is very essential in foretelling flights'on-time performance.

## I. INTRODUCTION

A flight delay is said to occur when an airline lands or takes off later than its scheduled arrival or departure time respectively. Conventionally if a flight's departure time or arrival time is greater than 15 minutes than its scheduled departure and arrival times respectively, then it is considered that there is a departure or arrival delay with respect to corresponding airports. Notable reasons for commercially scheduled flights to delay are adverse weather conditions, air traffic congestion, late reaching aircraft to be used for the flight from previous flight, maintenance and security issues [1]. American Airlines, Inc. (AA) is an American Airline that is based on Forth Worth, Texas. It is arguably world's largest airline on grounds of revenue, fleet size and scheduled passenger kilometres flown [2]. Hence, it is quite obvious for passengers to prefer American Airlines for domestic flights too. But flight delay concerned with American Airlines often seem sudden and unprecedented. Therefore, these delays make passengers lose their trust on such famous and internationally recognized airline. An Intelligent and Automated Prediction System is a must in this case that can predict possible airline delay. This model takes the flight details regarding American Airlines covering top 5 busiest airports of US, Hartsfield-Jackson Atlanta International Airport, Los Angeles International Airport, O'Hare International Airport, Dallas/Fort Worth International Airport and John F. Kennedy International Airport, as input, analyses it and gives the arrival prediction i.e., whether it will arrive at the concerned airport on-time (within 15 minutes after the scheduled arrival time) or not. This paper has been structured as an Introduction, Literature Review, Proposed Methodology, Training the Model, Implementation Details, Results and Conclusion.

## II. LITERATURE REVIEW

Many attempts have been by researchers in the past for predicting flight delays using Machine Learning, Deep Learning and Big Data approaches.

- Review on Machine Learning Techniques:
    - Chakrabarty et al. [3] proposed a Machine Learning Model using Gradient Boosting Classifier for predicting flight arrival delay of American Airlines covering 5 most busiest airports in US.
    - Manna et al. [4] explored and analyzed the flight data and developed a regression model using Gradient Boosting Regressor for predicting both Flight Departure and Arrival Delays respectively.
    - Choi et al. [5] applied Supervised Machine Learning Algorithms like decision tree, random forest, AdaBoost and k-Nearest Neighbours for predicting weather influenced flight delay.
    - Rebollo et al. [6] applied Random Forest on a air traffic network framework for predicting flight departure delays in future.
    - Ding et al. [7] proposed a Multiple Linear Regression approach for prediction of flight delay and also compared the model performance with Naive Bayes and C4.5 approaches.
    - Oza at al. [8] attempted weather induced flight delay prediction by implementing Weighted Multiple Linear Regression on weather-flight data having weather factors and weather delay probabilities.
    - Kalliguddi et al.[9] constructed regression models like Decision Tree Regressor, Random Forest Regressor and Multiple Linear Regressor on flight data for predicting both departure and arrival delays.
    - Ni et al.[10] implemented 6 supervised classification algorithms for the prediction of flight delay in US using Temporal and Geographical Information.

- Naul [11] applied Logistic Regression, Naive Bayes Classifier and Support Vector Machine on flight data for prediction of flight departure delay and also compared the performance between them.
- Review on Deep Learning Techniques:
  - Kim et al. [12] implemented a Deep Learning Approach using Recurrent Neural Networks (RNNs) for predicting flight delay.
  - Khanmohammadi et al. [13] proposed a Deep Learning Approach using Artifical Neural Network (ANN) and also introduced a new type of multilevel input layer ANN which is interpretable in the sense that the relationships between input and output variables can be easily chalked out and hence developing a flight delay prediction model for flights arriving or departing JFK Airport.
- Review in Big Data Approach(s):
  - Belcastro et al. [14] proposed a Big Data Approach by analyzing and mining flight information as well as corresponding weather conditions using parallel algorithms implemented as MapReduce programs executed on Cloud Platform for weather induced flight delay prediction.

### III. PROPOSED METHODOLOGY

#### A. The Dataset

*1) Desired Dataset:* The data for the study was collected from US Department of Transportation's Bureau of Transportation Statistics (BTS) [15]. Flight data was extracted from BTS in such a way that the dataset contained flight information regarding Year, Quarter, Month, Day of Month, Day of Week, Flight Number, Origin Airport ID, Origin Word Area Code, Destination Airport ID, Destination World Area Code, CRS Departure Time, Unique Airline ID (DOT ID Reporting Airline), CRS Arrival Time and the binary label, Arr Del 15 (it assumes two values, 0 and 1 where 0 means no Arrival Delay and 1 means Arrival Delay) for the years, 2015 and 2016.

*2) Data Collection and Analytics:* The following steps are followed to achieve the desired dataset:

- The aforementioned field names are checked with Filter Years and Filter Periods mentioned accordingly in the BTS Website. As periods are months, 24 datasets need to be downloaded to collect the data for the 2 years 2015 and 2016.
- These 24 datasets are concatenated row-wise to achieve the complete dataset.
- Now from the resulting dataset, the DOT_ID_Reporting_Airline is filtered to include only American Airlines by knowing the DOT ID Reporting Airline of American Airlines, which is found to be '19805'[16].
- The Origin Airport ID and Destination Airport ID are filtered to include the top 5 busiest airports in US where the Airport Codes (Origin and Destination Airport ID) are matched with the Airport Names using the Airport Codes Dataset available at Azure ML Studio's Sample Datasets.
- Finally, the attribute DOT ID Reporting Airline is removed from the dataset.

Hence, we get the desired dataset, required for the study. It contains 97,360 samples with 12 attributes and 1 label. Among 12 attributes, 2 attributes, CRS_Departure_Time and CRS_Arrival_Time are continuous while the remaining 10 attributes are categorical.

| ID | Attribute/Feature Name | Attribute Type |
|---|---|---|
| F1 | Year | Categorical |
| F2 | Quarter | Categorical |
| F3 | Month | Categorical |
| F4 | Day_of_Month | Categorical |
| F5 | Day_of_Week | Categorical |
| F6 | Flight_Num | Categorical |
| F7 | Origin_Airport_ID | Categorical |
| F8 | Origin_World_Area_Code | Categorical |
| F9 | Destination_Airport_ID | Categorical |
| F10 | Destination_World_Area_Code | Categorical |
| F11 | CRS_Departure_Time | Continuous |
| F12 | CRS_Arrival_Time | Continuous |

Table 1. Feature Study

#### B. Feature Selection

All the features in the flight dataset are highly significant and are tabulated separately in Table 1. Still some features are dropped from the dataset on several grounds:

- Year (F1) is a categorical feature having almost no variability as the dataset contains flight info of only 2 years, 2015 and 2016. So, the feature, F1 is dropped from the dataset.
- Quarter (F2) is somewhat, a repetitive feature as already the month of the flight is obtained from Month (F3) feature in the dataset. So F2 is dropped from the dataset. Also features like F8 and F10 seem as repetitive features as Origin and Destination Airports are already obtained from F7 and F9 respectively in the dataset. But apart from the dataset under consideration, there are instances where more than 1 airport may have the same World Area Code and so to preserve the versatility and consistency of our model, F8 and F10 are not dropped.

A correlation matrix is shown in Fig 1 in the form of a heatmap displaying Feature-to-Feature and Feature-to-Label Pearson Correlations where all the features are Continuous (F11 and F12) and the Arr Del 15 is the label.

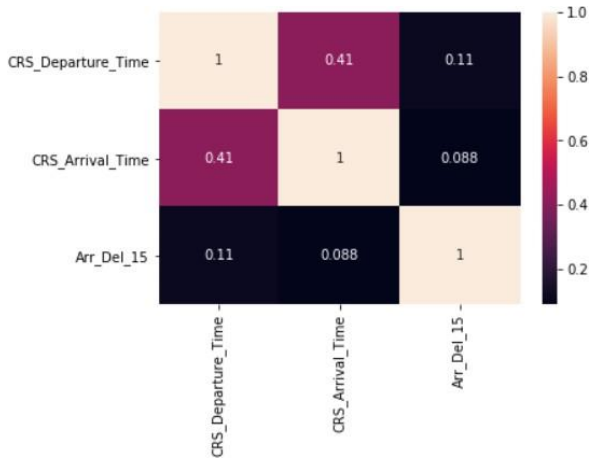

**Fig 1. Heat-Map showing Feature-to-Feature and Feature-to-Label's Pearson Correlation Coefficients**

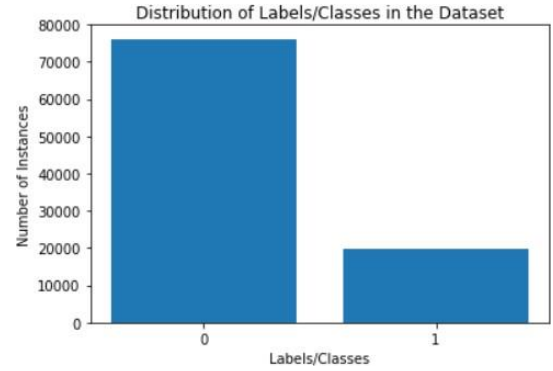

**Fig 2. Bar Visualization showing the imbalance between the 2 labels in the dataset**

*C. Data Preprocessing*

Before, training the model with the flight data, certain data pre-processing steps are essential. The pre-processing techniques employed here, are as follows:

*1) Handling Missing Values:* The dataset only contains missing values for the label, Arr Del 15. So, instances having missing labels are removed from the dataset.

*2) Label Encoding:* All the categorical features mentioned in Table 1 are label encoded, where alphabetically each category is assigned numbers beginning with 0.

*3) One-Hot Encoding:* This involves splitting of different categorical features into its own categories (distinct values assumed by them are the categories) where each and every category is assigned a binary value i.e., 0 if it does not belong to that category and 1 if it belongs to that category. This is done for the categorical features assuming lower number of distinct values or categories to avoid the curse of dimensionality. Here One-Hot Encoding is done for features, F7, F8, F9 and F10.

*4) Data Imbalance Removal or Data Balancing:* So far, in the implementation perspective, it is understandable that a binary classification needs to be performed on the label, Arr Del 15 which assumes binary values 0 and 1 where 0 indicates that there has been no Arrival Delay and 1 indicates that there has been Arrival Delay of the concerned flight. But the number of instances with label 0 have been 76090 and on the other hand, only 19668 samples of label 1 or flight delay instances are there. Hence, the dataset is highly imbalanced and the imbalance is visually shown in Fig 2 with the help of bars.

A technical solution to reduce the imbalance is Over-sampling. **Randomized-Synthetic Minority Over-sampling Technique (R-SMOTE)**: In this technique, the minority label i.e., label with lesser number of instances (here label 1) in the dataset is over-sampled. In other words, new artificial examples of the minority label are created to reduce the imbalance and match up with the majority label (here label 0) [17][18].

In Randomized-SMOTE:

- For each instance having minority label (here label 1), 2 other instances are randomly picked from the minority class excluding the considered instance.
- k number of data-points are linear interpolated between the 2 randomly selected instances such that,

$$k = (R - SMOTE\%)/100$$

- A data-point is linear interpolated between each of the linear interpolated data points in the previous step and the considered instance in the first step, resulting in k new data-points. These data-points are the Newly Sampled Instances obtained by Randomized SMOTE.

Here, a 200% Randomized-SMOTE is done on the dataset to reduce the imbalance between classes.

The reduction in the imbalance after 200% Over-Sampling is shown in Fig 3.

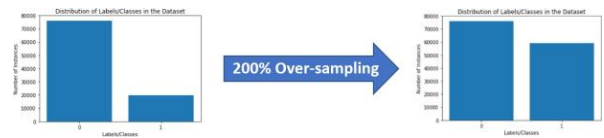

**Fig 3. Reduction in Data Imbalance after 200% R-SMOTE**

So, the resulting dataset has 1,35,094 instances among which 76,090 instances belong to label 0 and 59,004 instances belong to label 1.

*5) Shuffling and Splitting:* The whole dataset is shuffled in a consistent way and split into Training and Validation Sets with 80% of data constituting the Training Set and 20% of data constituting the Validation Set.

*6) Learning Algorithm:* The Machine Learning Algorithm used to construct the Binary Prediction Model is an Ensemble

Tree Boosting Algorithm known as Gradient Boosting Classifier. The intuitive idea regarding the working mechanism of the classifier is that, it forms a series of Decision Trees in which every next tree corrects the error caused by the tree preceding it. The Algorithm of the Gradient Boosting Classifier is given below as Algorithm 1:

- Input: training set $Z=\{(x_1, y_1), ., (x_n, y_n)\}$
- M: number of iterations
- v: learning rate

1. $f_0(x)=log(p_1/(1-p_1))$
2. For m=1...M:
3.   $g_i=dL(y_i, f_m(x_i))/df_m(x_i)$
4.   A tree, $h_m(x_i)$ is fit to target $g_i$
5.   $p_m=argmax_p Q[f_{m-1}(x)+p*h_m(x)]$
6.   $f_m(x)=f_{m-1}(x)+v*p_m*h_m(x_i)$
7. return $f_M(x)$

**Algorithm 1: Gradient Boosting Classifier**

The Data Pre-Processing is done in 2 ways:

- Strategy 1: The Data Pre-Processing is done by skipping the Data Imbalance Removal step discussed in Section III.C.4
- Strategy 2: The Data Pre-Processing is done by following all the steps as mentioned as above.

## IV. TRAINING THE MODEL

The Gradient Boosting Classifier Model is tuned with Grid Search for obtaining the best set of hyper-parameters following both Strategy 1 and Strategy 2. After training the model with Grid Search applied on Gradient Boosting Classifier, 300 estimators and maximum depth of 5 are obtained as the best set of hyper-parameters in Strategy 1 and 400 estimators with maximum depth of 5 in Strategy 2. The summary of Grid-Search Tuning of the Model following Strategy 1 and Strategy 2, based on Mean Score, is shown in Fig 4 and Fig 5 respectively.

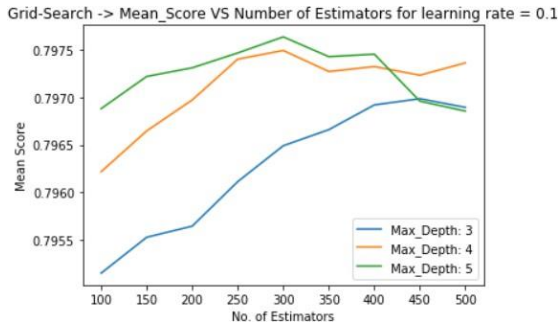

**Fig 4. Grid Search Summary on Mean Score following Strategy 1**

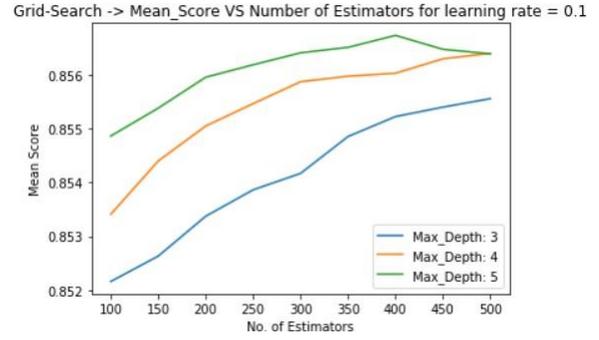

**Fig 5. Grid Search Summary on Mean Score following Strategy 2**

## V. IMPLEMENTATION DETAILS

The Data Preprocessing (**excluding Section III.C.4**) and Model Development are done using Python's Scikit-Learn Machine Learning Toolbox on a machine with Intel(R) Core(TM) i5-8250U processor, CPU @ 1.60 GHz 1.80 GHz and 8 GB RAM.

## VI. RESULTS

The metrics on which model performance is evaluated are as follows:

- Training Accuracy describes how correctly, the samples in the Training Set are predicted by the model.
- Validation Accuracy describes how correctly, the model predicts the samples in the Validation Set.
- Sensitivity or Recall is defined as the fraction of correctly identified positives.

$$Recall = TP/TP + FN$$

- Precision is defined as the proportion of correctly predicted positive observations of the total predicted positive observations.

$$Precision = TP/TP + FP$$

- F1-Score is the harmonic mean (HM) of Precision and Recall.
- Area Under Receiver Operator Characteristic Curve (AU-ROC): ROC Curve is the plot of True Positive Rate vs False Positive Rate. An Area under ROC Curve of greater than 0.5, is acceptable.
- The structure of Confusion Matrix is shown in Fig 6.

|  |  | Predicted class | |
|---|---|---|---|
|  |  | Class = Yes | Class = No |
| Actual Class | Class = Yes | True Positive | False Negative |
|  | Class = No | False Positive | True Negative |

**Fig 6. Structure of Confusion Matrix**

All the results obtained from Strategy 1 and Strategy 2 are shown in Table 2.

| Strategies | Training Accuracy | Validation Accuracy | Recall | Precision | F1-Score | AUROC |
|---|---|---|---|---|---|---|
| Strategy 1 | 80.89% | 80.18% | 0.80 | 0.77 | 0.74 | 0.71 |
| Strategy 2 | 86.68% | 85.73% | 0.86 | 0.88 | 0.85 | 0.90 |

**Table 2. Results and Findings**

The Confusion Metrics for Strategy 1 and Strategy 2 are given in Fig 7 and Fig 8 respectively.

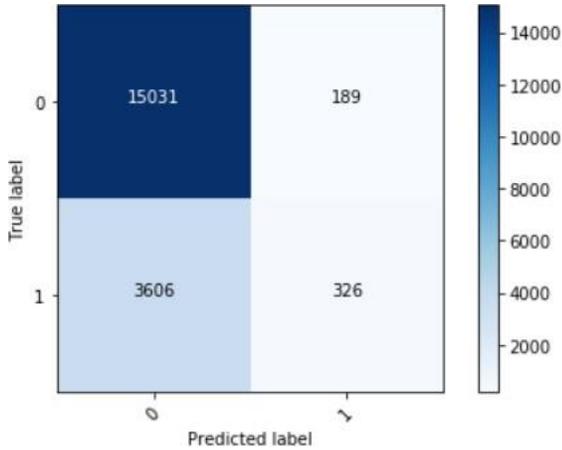

**Fig 7. Confusion Matrix for Strategy 1**

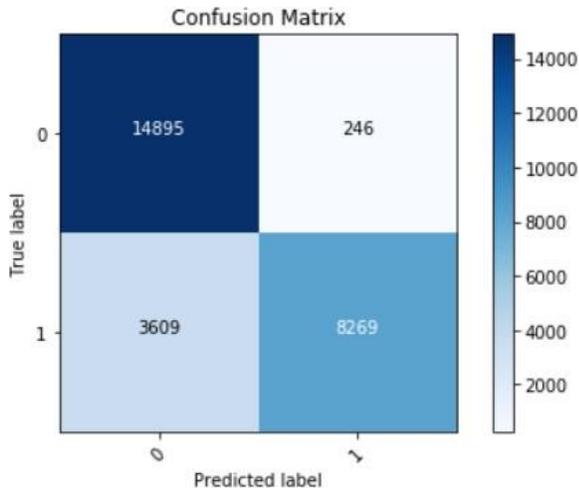

**Fig 8. Confusion Matrix for Strategy 2**

The Receiver Operator Characteristic Curve (ROC Curve) is implemented using the decision function attribute of Gradient Boosting Classifier which returns values that give a measure of how far a data-point is away from the Decision Boundary from either side (negative value for opposite side). The ROC Curves for Strategy 1 and Strategy 2 are shown in Fig 9 and Fig 10 respectively.

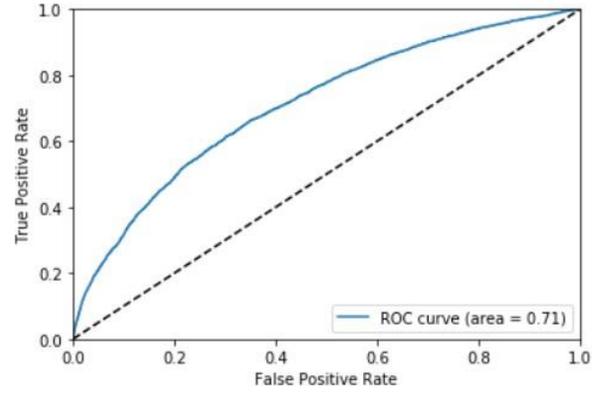

**Fig 9. ROC Curve showing the Area Under the Curve for Strategy 1**

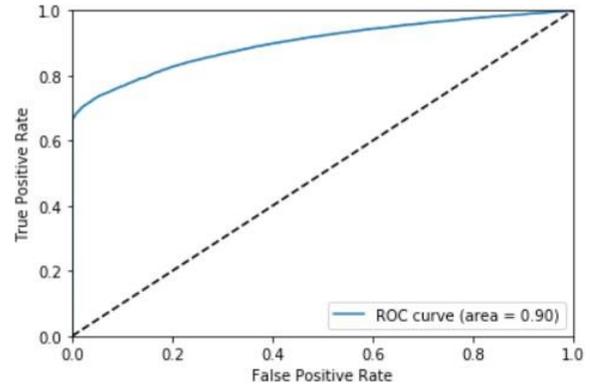

**Fig 10. ROC Curve showing the Area Under the Curve for Strategy 2**

- According to the obtained Training Accuracy and Validation Accuracy of both Strategy 1 and Strategy 2, it can be concluded that the 2 models are absolute perfect fits.
- The Area Under the ROC Curve (AUROC) shown in Fig 9 and Fig 10 for Strategy 1 and Strategy 2 are 0.71 and 0.90 respectively. Both are highly acceptable as more the AUROC (towards 1.0), better the performance of the model.

Now, comparing the performances of Strategy 1 and Strategy 2, it is evident that Over-sampling any Imbalanced Dataset and hence, reducing its degree of imbalance, helps improving and boosting model performance to a much greater extent.

A direct comparison is shown in Table 3 between 3 models as they are developed using the same dataset:

- Chakrabarty et al.[3]
- Model constructed by Strategy 1
- Model constructed by Strategy 2

on the following basis:

- Proposed Methodology
- Learning Algorithm Used
- Model Performance excluding AUROC as the ROC curve, implemented by Chakrabarty et al. [3] is based on thresh-holding which is done directly by the predictions of the label, (Arr Del 15) returned by the model and hence, penalizing the Area Under the Curve.

| Comparison Parameters | | Chakrabarty et al. [3] | Strategy 1 | Strategy 2 |
|---|---|---|---|---|
| Proposed Methodology | Feature Selection | - | F1 and F2 are dropped | F1 and F2 are dropped |
| | Data Preprocessing — Handling Missing Values | Samples containing missing values are removed. | Samples containing missing values are removed. | Samples containing missing values are removed. |
| | Data Preprocessing — Label Encoding | Categorical Features are coded alphabetically starting from 0. | Categorical Features are coded alphabetically starting from 0. | Categorical Features are coded alphabetically starting from 0. |
| | Data Preprocessing — One-Hot-Encoding | - | F7, F8, F9 and F10 are One-Hot-Encoded | F7, F8, F9 and F10 are One-Hot-Encoded |
| | Data Imbalance Removal | - | - | 200% RANDOM_SMOTE |
| Learning Algorithm Used and Model Training | | Gradient Boosting Classifier | Gradient Boosting Classifier (Grid-Search) | Gradient Boosting Classifier (Grid-Search) |
| Model Performance | Training Accuracy | 81.06% | 80.89% | 86.68% |
| | Validation Accuracy | 79.72% | 80.18% | 85.73% |
| | Recall | 0.80 | 0.80 | 0.86 |
| | Precision | 0.76 | 0.77 | 0.88 |
| | F1-Score | 0.74 | 0.74 | 0.85 |

Table 3. Grand Comparison Chart between 3 models

## VII. Conclusion

This paper proposed a hyper-parameter tuned approach by the application of Grid Search on Gradient Boosting Classifier Model on flight data. Also here, Over-Sampling technique, Randomized SMOTE is applied for Data Balancing, in which the resulting performance boosting has also been shown. Finally, the Validation Accuracy, obtained is 85.73% which is, by the best of our knowledge, has been the best ever numeric accuracy clocked by any Flight Delay Prediction Model on this dataset.

Future Scope of this work involves application of more advanced and novel pre-processing techniques, sampling algorithms and Machine Learning-Deep Learning Hybrid Models tuned with Grid Search for achieving better model performance.